\documentclass[10pt, a4paper]{article}
\usepackage{lrec}
\usepackage{multibib}
\newcites{languageresource}{Language Resources}
\usepackage{graphicx}
\usepackage{tabularx}
\usepackage{soul}
 \usepackage{multirow}
\usepackage{epstopdf}
\usepackage{hyperref}
\usepackage{xstring}
\usepackage{color}
\usepackage{arabtex}
\usepackage{utf8}
\usepackage{tipa}

\newcommand{\sys}{{AraNet }}

\title{AraNet: A Deep Learning Toolkit for Arabic Social Media}
\name{Muhammad Abdul-Mageed, Chiyu Zhang, Azadeh Hashemi, El Moatez Billah Nagoudi}

\address{Natural Language Processing Lab, University of British Columbia\\
        \{muhammad.mageeed, azadeh.hashemi, moatez.nagoudi\}@ubc.ca\\chiyuzh@mail.ubc.ca}

\date{}

\abstract{
We describe AraNet, a collection of deep learning Arabic social media processing tools. Namely, we exploit an extensive host of both publicly available and novel social media datasets to train bidirectional encoders from transformers (BERT) focused at social meaning extraction. AraNet models predict age, dialect, gender, emotion, irony, and sentiment. AraNet delivers state-of-the-art performance on a number of these tasks and performs competitively on others. \sys is exclusively based on a deep learning framework, giving it the advantage of being feature-engineering free. To the best of our knowledge, \sys is the first to performs predictions across such a wide range of tasks for Arabic NLP. As such, \sys has the potential to meet critical needs. We publicly release \sys to accelerate research, and to facilitate model-based comparisons across the different tasks.}

\begin{document}
\maketitleabstract

\setcode{utf8}
\setarab
\novocalize

\section{Introduction}
\begin{figure}[h]
\begin{centering}
 \frame{\includegraphics[width=\linewidth]{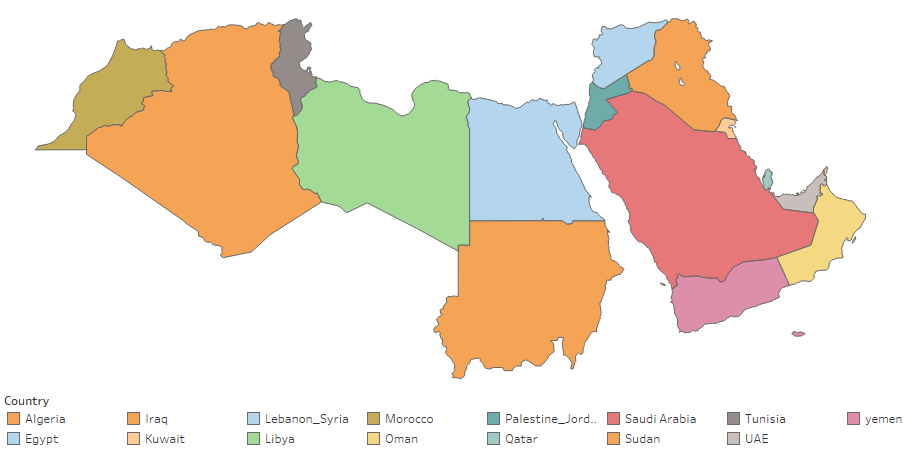}}
  \caption{A map of Arab countries. Our different datasets cover varying regions of the Arab world as we describe in each section.}
  \label{fig:ara_w_wiki}
  \end{centering}
\end{figure}

The proliferation of social media has made it possible to study large online communities at scale. This offers opportunities to make important discoveries, facilitate decision making, guide policies, improve health and well-being, aid disaster response, attend to population needs in pandemics such as the current COVID-19, etc. The wide host of languages, languages varieties, and dialects used on social media and the nuanced differences between users of various backgrounds (e.g., different age groups, gender identities) make it especially difficult to derive sufficiently valuable insights based on single prediction tasks. For these reasons, it is highly desirable to develop natural  language processing (NLP) tools that can help piece together more complete pictures of events impacting individuals of different identities across different geographic regions. In this work, we propose \sys, a suit of tools that has the promise to play such a role of Arabic social media processing. 
\paragraph{}
For Arabic, a collection of languages and varieties spoken by a wide population of $\sim 400$ million native speakers covering a vast geographical region (shown in Figure~\ref{fig:ara_w_wiki}), no such suite of tools currently exists. Many works have focused on sentiment analysis, e.g., ~\cite{mageed2014samar,nabil2015astd,elsahar2015building,al2015deep,al2018arabic,al2019using,al2019comprehensive,farha2019mazajak} and dialect identification~\cite{elfardy2013sentence,zaidan2011arabic,zaidan2014arabic,cotterell2014multi,zhang2019no,bouamor2019madar}. However, there is rarity of tools for other tasks such as gender and age detection. This motivates our toolkit, which we hope can meet the current critical need for studying Arabic communities online. This is especially valuable given the waves of protests, uprisings, and revolutions that have swept the region during the last decade.

\begin{figure}[h]
\begin{centering}
 \frame{\includegraphics[width=\linewidth]{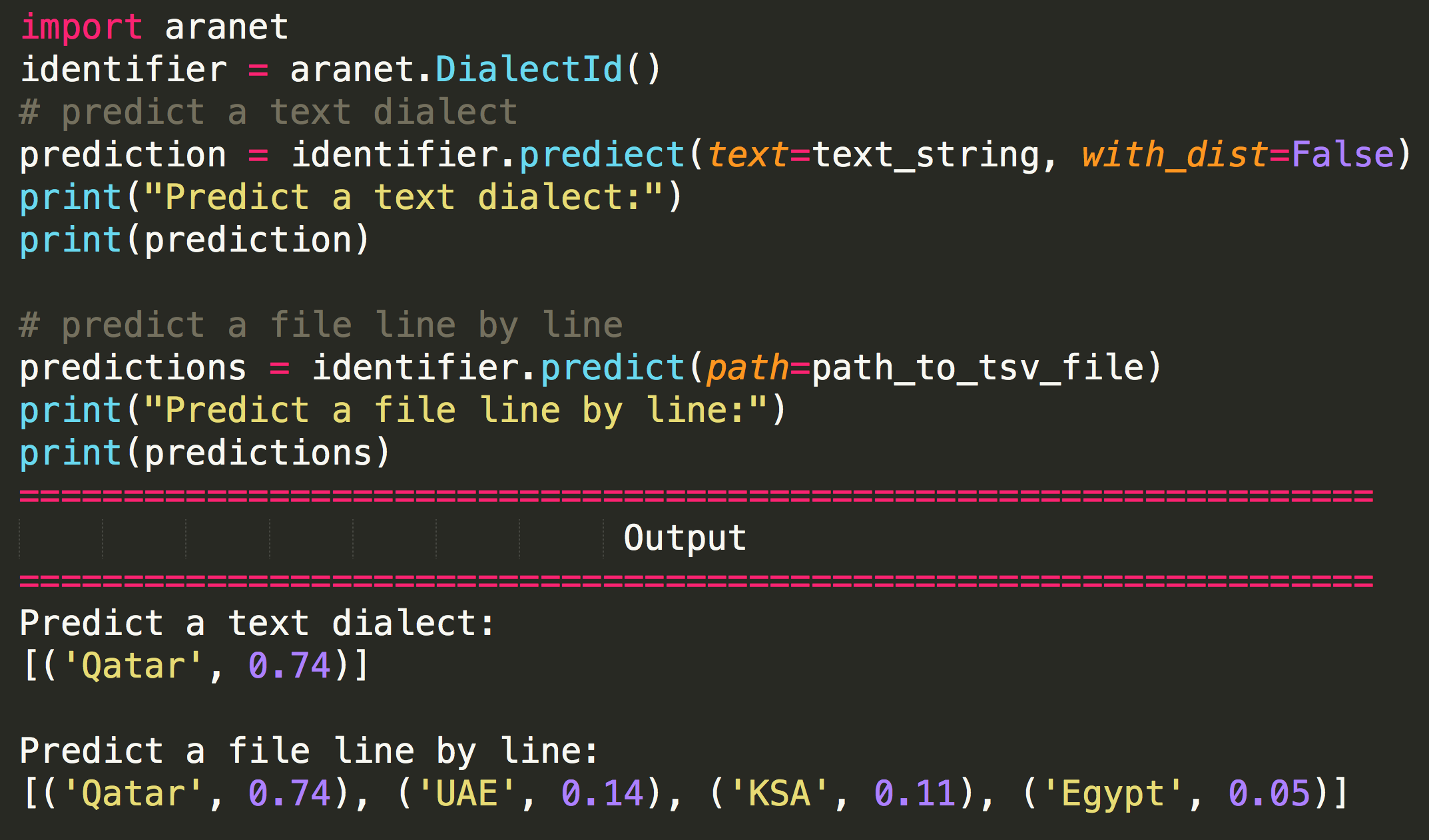}}
  \caption{AraNet usage and output as a Python library.}
  \label{fig:uasge1}
  \end{centering}
\end{figure}

\paragraph{}
Although we create new models for tasks such as sentiment analysis and gender detection as part of AraNet, our primary focus is to provide strong baselines across the various tasks. We believe this will facilitate comparisons across models. This is particularly useful due to absence of standardization across datasets for many of the tasks, and given the somewhat ephemeral nature of parts of some types of these data. In particular, many tasks are developed based on social media posts such as tweets that are distributed under restrictive conditions. For example, Twitter terms require release of data only in the form of tweet ids, making it challenging to acquire $100\%$ of these tweets especially once the data are several months old. These reasons make \textit{model-based comparisons} appealing, as a way to measure research progress in absence of easy benchmarking. Our general approach is to adopt sensible baselines across the various AraNet tasks, but we do not necessarily explicitly compare to all previous research. This is the case since most existing works either exploit \textit{smaller data} (and so it will not be a fair comparison), use \textit{methods pre-dating BERT} (and so will likely be outperformed by our models). In addition, we note  that although it would have been possible to acquire better results by feature engineering (especially on smaller datasets), our main goal is to keep our models free of laborious feature engineering. In some tasks, we even acquire better results that what is reported here by adopting more involved methods. But, again, we do our best here to keep all models relatively comparable (and as simple as possible) in terms of the methods employed to acquire them. Our hope is that, by adopting model-based comparisons, we can help accelerate progress on Arabic social media processing. For these reasons, we also package models from our recent works on dialect~\cite{zhang2019no} and irony~\cite{zhang2019irony} as part of AraNet.

\paragraph{}

The rest of the paper is organized as follows: In Section~\ref{sec:methods} we describe our methods. In Section~\ref{sec:data_models}, we describe or refer to published literature for the data we exploit for each task. Also in Section~\ref{sec:data_models}, we provide results from our models. Section~\ref{sec:design_use} is about \sys design and use, and Section~\ref{sec:ethics} is about ethical considerations. We overview related works in Section~\ref{sec:rel_works}, and conclude in Section~\ref{sec:conc}.

\section{Methods}~\label{sec:methods}
\subsection{Supervised BERT}

\textbf{Transformer.} Across all our tasks, we use Bidirectional Encoder Representations from Transformers (BERT). BERT is based on the Transformer architecture of ~\cite{vaswani2017attention}, which we briefly introduce here. The Transformer depends solely on self-attention, thus allowing for parallelizing the network (unlike RNNs). It is an encoder-decoder architecture where the \textit{encoder} takes a sequence of symbol representations $x^{(i)} \dots x^{(n)}$, maps them into a sequence of continuous representations $z^{(i)} \dots x^{(n)}$ that are then used by the \textit{decoder} to generate an output sequence $y^{(i)} \dots y^{(n)}$, one symbol at a time. This is performed using \textit{self-attention}, where different positions of a single sequence are related to one another. The Transformer employs an attention mechanism based on a function that operates on \textit{queries}, \textit{keys}, and \textit{values}. The attention function maps a query and a set of key-value pairs to an output, where the output is a weighted sum of the values. For each value, a weight is computed as a compatibility function of the query with the corresponding key. This particular version of attention is a scaled dot product of queries and keys (each of $d_k$) that is scaled by a factor of $\frac{1}{\sqrt{d_k}}$ on which a softmax is applied to acquire the weights on the values. The scaled dot product attention is computed as as a set of queries, keys, and values in three matrices Q, K, and V, respectively, follows:

\begin{equation}
 Attention \left(\textit{Q, K, V}\right) = softmax \left( \frac{QK^T}{\sqrt{d_k}}\right)V  \\ 
\end{equation}

Encoder of the Transformer in~\cite{vaswani2017attention} has 6 attention layers, each of which has \textit{h} attention heads (\textit{multi-head attention}) to allow the model to jointly attend to information from different representation subspaces across different positions. Each of the 6 layers also has a simple, fully-connected feed-forward network (FFN) that is applied to each position separately and identically that different parameters across the different layers. Decoder of the Transformer is similar to the encoder but has a third sub-layer that performs multi-head attention over the encoder stack. Since the Transformer discards with both recurrence and convolution, it resorts to the so-called \textit{positional encoding} (based on sin and cosine functions) at the bottoms of the encoder and decoder stacks as a way to capture order of the sequence. We now introduce BERT.

\paragraph{}

\textbf{BERT.} BERT involves two self-supervised learning tasks, (1) \textit{masked language models (Masked LM)} and (2) \textit{next sentence prediction}. Since BERT uses bidirectional conditioning, a given percentage of random input tokens are masked and the model attempts to predict these masked tokens. This is the Masked LM task, where masked tokens are simply replaced by a string \texttt{[MASK]} .~\cite{devlin2018bert} mask 15\% of the tokens (the authors use WordPieces) and feed the final hidden vectors of these masked tokens to an output softmax over the vocabulary. The next sentence prediction task is just binary classification. For a given sentence \texttt{S}, two sentences \texttt{A} and \texttt{B} are generated where \texttt{A} (positive class) is an actual sentence from the corpus and \texttt{B} is a randomly chosen sentence (negative class). Once trained on an unlabeled dataset, BERT can then be fine-tuned with supervised data for a downstream task (e.g., text classification, question answering).

\paragraph{}

All our models are trained in a fully supervised fashion, with dialect id being the only task where we leverage semi-supervised learning. We briefly outline our semi-supervised methods next.

\subsection{Self-Training} 

Only for the dialect id task, we investigate augmenting our human-labeled training data with automatically-predicted data from self-training. Self-training is a wrapper method for semi-supervised learning~\cite{Triguero2015,pavlinek2017text} where a classifier is initially trained on a (usually small) set of labeled samples $\textbf{\textit{D}}^{l}$, then is used to classify an unlabeled sample set $\textbf{\textit{D}}^{u}$. Most confident predictions acquired by the original supervised model are added to the labeled set, and the model is iteratively re-trained. We perform self-training using different confidence thresholds and choose different percentages from predicted data to add to our dialect training set. We only report best settings here, and the reader is referred to our winning system on the MADAR shared task for more details on these different settings~\cite{zhang2019no}.

\subsection{Implementation \& Models Parameters} 

For all our tasks, we use the BERT-Base Multilingual Cased model released by the authors~\footnote{\url{https://github.com/google-research/bert/blob/master/multilingual.md}.}. The model is trained on 104 languages (including Arabic) with 12 layer, 768 hidden units each, 12 attention heads, and has 110M parameters in entire model. The model has 119,547 shared WordPieces vocabulary, and was pre-trained on the entire Wikipedia for each language. For fine-tuning, we use a maximum sequence size of 50 tokens and a batch size of 32. We set the learning rate to $2e-5$ and train for 15 epochs~\footnote{For dialect id, we trained only for 10 epochs. This was based on monitoring loss on a development set.} and choose the best model based on performance on a development set. We use the same hyper-parameters in all of our BERT models. We fine-tune BERT on each respective labeled dataset for each task. For BERT input, we apply WordPiece tokenization, setting the maximal sequence length to 50 words/WordPieces. For all tasks, we use a TensorFlow implementation. An exception is the sentiment analysis task, where we used a PyTorch implementation with the same hyper-parameters but with a learning rate $2e-6$.~\footnote{We find this learning rate to work better when we use PyTorch.}

\paragraph{}
\textbf{Pre-processing.} Most of our training data in all tasks come from Twitter. Exceptions are in some of the datasets we use for sentiment analysis, which we point out in Section~\ref{subsec:senti}. Our pre-processing thus incorporates methods to clean tweets, other datasets (e.g., from the news domain) being much less noisy. For pre-processing, we remove all usernames, URLs, and diacritics in the data. 

\section{Data and Models}\label{sec:data_models}

\subsection{Age and Gender}
\begin{table*}[]
\centering\begin{tabular}{l|l|l|l|l|l|l|l}
\hline
\multirow{2}{*}{\textbf{\textbf{Data split}}} & \multicolumn{2}{c|}{\textbf{Under 25}} & \multicolumn{2}{c|}{\textbf{25 until 34}} & \multicolumn{2}{c|}{\textbf{35 and up}} & \multicolumn{1}{c}{\multirow{2}{*}{\textbf{\# of tweets}}} \\ \cline{2-7}
                                              & \textbf{Female}     & \textbf{Male}    & \textbf{Female}      & \textbf{Male}      & \textbf{Female}     & \textbf{Male}     & \multicolumn{1}{c}{}                                  \\ \hline
\textbf{TRAIN}                                & 215,950             & 213,249          & 207,184              & 248,769            & 174,511             & 226,132           & 1,285,795                                              \\ 
\textbf{DEV}                                  & 27,076              & 26,551           & 25,750               & 31,111             & 21,942              & 28,294            & 160,724                                                \\ 
\textbf{TEST}                                 & 26,878              & 26,422           & 25,905               & 31,211             & 21,991              & 28,318            & 160,725                                                \\ \hline
\textbf{ALL}                                  & 269,904             & 266,222          & 258,839              & 311,091            & 218,444             & 282,744           & 1,607,244                                              \\ \hline
\end{tabular}\caption{Distribution of age and gender classes in our Arab-Tweet data splits}
\label{tab:classes}
\end{table*}

\textbf{Arab-Tweet.} For modeling age and gender, we use Arap-Tweet~\cite{zaghouani2018arap}~\footnote{The resource is an Arabic \textbf{p}rofiling dataset, and hence the sequence ``Arap" with an ``p".}, which we will refer to as \textit{Arab-Tweet}. Arab-tweet comprises 11 Arabic regions from 17 different countries.~\footnote{Counts are based on the distribution we received from the authors.} For each region, data from 100 Twitter users were crawled. Users needed to have posted at least 2,000 tweets and were selected based on an initial list of seed words characteristic of each region. The seed list included words such as <برشة> /barsha/ ‘many’ for Tunisian Arabic and  <وايد> /wayed/ ‘many’ for Gulf Arabic.~\cite{zaghouani2018arap} employed human annotators to verify that users do belong to each respective region. Annotators also assigned gender labels from the set \textit{{male, female}} and age group labels from the set \textit{{under-25, 25-to34, above-35}} at the user-level, which in turn is the tag for tweet level. Tweets with less than 3 words and re-tweets were removed. Refer to~\cite{zaghouani2018arap} for details about how annotation was carried out. We provide a description of the data in Table~\ref{tab:classes}. Table~\ref{tab:classes} also provides class breakdown across our splits. We note that~\cite{zaghouani2018arap} do not report classification models exploiting the data. Although age and gender are user-level tasks, note that we train \textit{tweet-level} age and gender models. However, tweet-level models can easily be ported to user-level by simply taking the majority class based on softmax-thresholding as we show in~\cite{zhang2019no}.~\footnote{Arab-Tweet is also distribute only with tweet-level labels (i.e., without user ids), thus making it not possible to model age and gender at the user level exploiting the data.} 
\paragraph{}
We shuffle the Arab-tweet dataset and split it into 80\% training (TRAIN), 10\% development (DEV), and 10\% test (TEST). The distribution of classes in our splits is in Table~\ref{tab:classes}. For pre-processing, we reduce 2 or more consecutive repetitions of the same character into only 2 and remove diacritics. With this dataset, we train a small unidirectional GRU (\textit{small-GRU}) with a single 500-units hidden layer and dropout$=0.5$ as a baseline. Small-GRU is trained with the TRAIN set, batch size = 8, and up to 30 words of each sequence. Each word in the input sequence is represented as a trainable 300-dimension vector. We use the top 100K words from TRAIN which are weighted by mutual information as our vocabulary in the embedding layer. We evaluate the model on the blind TEST set. Table~\ref{tab:arab-tweet} shows that small-GRU obtains 36.29\% acc. on age classification, and 53.37\% acc. on gender detection.  Table~\ref{tab:arab-tweet} also shows performance of the fine-tuned BERT model. BERT significantly outperforms our baseline on the two tasks. It improves 15.13\% acc. (for age) and 11.93\% acc. (for gender) over the small-GRU.

\begin{table}[]
\centering
\begin{tabular}{l|l|c|c|l}
\hline
\multirow{2}{*}{\textbf{}} & \multicolumn{2}{c|}{\textbf{Age}}                 & \multicolumn{2}{c}{\textbf{Gender}}              \\ \cline{2-5} 
                           & \multicolumn{1}{c}{\textbf{DEV}} & \textbf{TEST} & \textbf{DEV} & \multicolumn{1}{c}{\textbf{TEST}} \\ \hline
\textbf{small-GRU}         & 36.13                             & 36.29         & 53.39        & 53.37                              \\ 
\textbf{BERT}              & \textbf{50.95}                             & \textbf{51.42}         & \textbf{65.31}        & \textbf{65.30}                              \\ \hline
\end{tabular}\caption{Model performance in accuracy of Arab-Tweet age and gender classification tasks. } \label{tab:arab-tweet}
\end{table}

\paragraph{}
\textbf{UBC Twitter Gender Dataset.}
We also develop an in-house Twitter dataset for gender. We manually labeled 1,989 users from each of the 21 Arab countries. The data had 1,246 ``male", 528 ``female", and 215 unknown users. We remove the ``unknown" category and balance the dataset to have 528 from each of the two ``male" and ``female" categories. We ended with 69,509 tweets for ``male" and 67,511 tweets for ``female". We split the users into 80\% TRAIN (110,750 tweets for 845 users), 10\% DEV (14,158 tweets for 106 users), and 10\% TEST (12,112 tweets for 105 users). We then model this dataset with BERT and evaluate on DEV and TEST. Table~\ref{tab:gender_ubc} shows that fine-tuned model obtains 62.42\% acc. on DEV and 60.54\% acc. on TEST. These results are 2.89\% and 4.76\% less than performance on Arab-Tweet, perhaps reflecting more diversity in UBC-Gender data which also makes it more challenging. Another potential reason for this accuracy drop could be that, for this tweet-level task, some tweets from the same user occur across our TRAIN/DEV/TEST splits. This was unavoidable since Arab-Tweet is distributed without user ids, thus not making it possible for us to prevent user-level data leakage into the two tweet-level classification tasks of age and gender we report here. We alleviate this issue for gender by annotating and developing on UBC-Gender where we control for user-level data distribution across the splits as explained earlier. 
\paragraph{}
We also combine the Arab-tweet gender dataset with our UBC-Gender dataset for gender on training, development, and test, respectively, to obtain new TRAIN, DEV, and TEST. We fine-tune BERT on the combined TRAIN and evaluate on combined DEV and TEST. As Table~\ref{tab:gender_ubc} shows, the model obtains 65.32\% acc. on combined DEV, and 65.32\% acc. on combined TEST. This is the model we package in AraNet.
\begin{table}[]
\centering
\begin{tabular}{l|l|l}
\hline
\textbf{}              & \multicolumn{1}{c|}{\textbf{DEV}} & \multicolumn{1}{c}{\textbf{TEST}} \\ \hline
\textbf{UBC\_TW\_Gender} & 62.42                            & 60.54                             \\ 
\textbf{Gender\_comb}    & \textbf{65.32}                            & \textbf{65.32}                             \\ \hline
\end{tabular}\caption{Model performance in accuracy. UBC\_TW\_Gender refers to the model trained on UBC Twitter Gender dataset. Gender\_Comb denotes the model trained on the Arab-Tweet and UBC-Gender combined TRAIN data split. Each model is evaluated on the corresponding DEV and TEST sets.}~\label{tab:gender_ubc}
\end{table}

\subsection{Dialect}

The dialect identification model in \sys is based on our winning system in the MADAR shared task 2 \cite{bouamor-etal-2019-madar} as described in~\cite{zhang2019no}. The corpus is divided into training, development, and test; and the organizers masked test set labels. We lost some tweets from TRAIN when we crawled using tweet ids, ultimately acquiring 2,036 (TRAIN-A), 281 (DEV) and 466 (TEST). We also make use of the task 1 corpus (95,000 sentences~\cite{bouamor2018madar}). More specifically, we concatenate the task 1 data to the training data of task 2, to create TRAIN-B. Again, note that TEST labels were only released to participants after the official task evaluation. Table ~\ref{tab:data} shows statistics of the data. More information about the data is in ~\cite{bouamor2018madar}. We use TRAIN-A to perform supervised modeling with BERT and TRAIN-B for self training, under various conditions. We refer the reader to~\cite{zhang2019no} for more information about our different experimental settings on dialect id. We acquire our best results with self-training, with a classification accuracy of 49.39\% and $F_1$ score at 35.44. This is the winning system model in the MADAR shared task and we showed in~\cite{zhang2019no} that our tweet-level predictions can be ported to user-level prediction. On user-level detection, our models perform superbly, with 77.40\% acc. and 71.70\% $F_1$ score on unseen MADAR TEST.

\begin{table}[h!]
\centering
\begin{tabular}{l|l|c|c}
\hline
\multirow{2}{*}{\textbf{}} & \multicolumn{3}{c}{\textbf{\# of tweets}}    \\ \cline{2-4} 
                           & \textbf{TRAIN} & \textbf{DEV} & \textbf{TEST} \\ \hline
\textbf{TRAIN-A}           & 193,086        & 26,588       & 43,909        \\ 
\textbf{TRAIN-B}           & 288,086        & --           & --            \\ \hline
\end{tabular}
\caption{Distribution of classes within the MADAR twitter corpus.}\label{tab:data}
\end{table}


\subsection{Emotion}

We make use of two datasets, LAMA-DINA and LAMA-DIST~\cite{alhuzali2018enabling}. The LAMA-DINA dataset is a Twitter dataset with a combination of gold labels from~\cite{abdul2016dina} and distant supervision labels. The tweets are labeled with the Plutchik $8$ primary emotions from the set: \textit{\{anger, anticipation, disgust, fear, joy, sadness, surprise, trust\}}. The distant supervision approach depends on use of seed phrases with the Arabic first person pronoun \<انا>  (Eng. ``I") + a seed word expressing an emotion, e.g., \<فرحان> (Eng. ``happy"). The manually labeled part of the data comprises tweets carrying the seed phrases verified by human annotators $9,064$ tweets for inclusion of the respective emotion. LAMA-DIST ($182,605$ tweets)~\footnote{These statistics are based on minor cleaning of the data to remove short tweets $< 3$ words and residuals of the seeds used for collecting the data.} is only labeled using distant supervision. For more information about the dataset, readers are referred to~\cite{alhuzali2018enabling}. The data distribution over the emotion classes is in Table~\ref{tab:emo_data}. We combine LAMA+DINA and LAMA-DIST training set and refer to this new training set as LAMA-D2 ($189,903$ tweets). We fine-tune BERT on the LAMA-D2 and evaluate the model with same DEV and TEST sets from LAMA+DINA. On DEV set, the fine-tuned BERT model obtains 61.43\% acc. and 58.83 $F_1$. On TEST set, we acquire 62.38\% acc.	and 60.32\% $F_1$.

\begin{table}[h]
\centering
\begin{tabular}{l|r|r|r|r}
\hline
\multicolumn{1}{c|}{\multirow{2}{*}{\textbf{}}} & \multicolumn{2}{c|}{\textbf{LAMA+DINA}}                             & \multicolumn{2}{c}{\textbf{LAMA-DIST}}                             \\ \cline{2-5} 
\multicolumn{1}{c|}{}                           & \multicolumn{1}{c|}{\textbf{\#}} & \multicolumn{1}{c|}{\textbf{\%}} & \multicolumn{1}{c|}{\textbf{\#}} & \multicolumn{1}{c}{\textbf{\%}} \\ \hline
\textbf{anger}                                   & 1,038                           & 11.45                            & 3,650                           & 2.00                             \\ 
\textbf{anticipation}                            & 933                             & 10.29                            & 24,672                          & 13.51                            \\ 
\textbf{disgust}                                 & 1,069                           & 11.79                            & 2,478                           & 1.36                             \\ 
\textbf{fear}                                    & 1,434                           & 15.82                            & 28,315                          & 15.51                            \\ 
\textbf{happy}                                   & 1,364                           & 15.05                            & 55,253                          & 30.26                            \\ 
\textbf{sad}                                     & 1,195                           & 13.18                            & 27,584                          & 15.11                            \\ 
\textbf{surprise}                                & 1,167                           & 12.88                            & 15,106                          & 8.27                             \\ 
\textbf{trust}                                   & 864                             & 9.53                             & 25,547                          & 13.99                            \\ \hline
\textbf{total}                                   & 9,064                           & 100.00                           & 182,605                         & 100.00                           \\ \hline
\end{tabular}
\caption{Emotion class distribution in LAMA+DINA and LAMA-DIST datasets.}
\label{tab:emo_data}
\end{table}


\subsection{Irony}
We use the dataset for irony identification on Arabic tweets released by IDAT@FIRE2019 shared task~\cite{idat2019}. The shared task dataset contains $5,030$ tweets related to different political issues and events in the Middle East taking place between 2011 and 2018. Tweets are collected using pre-defined keywords (i.e., targeted political figures or events) and the positive class involves ironic hashtags such as \#sokhria, \#tahakoum, and \#maskhara (Arabic variants for ``irony"). Duplicates, retweets, and non-intelligible tweets are removed by organizers. Tweets involve both MSA as well as dialects at various degrees of granularity such as \textit{Egyptian}, \textit{Gulf}, and \textit{Levantine}. 

IDAT@FIRE2019 \cite{idat2019} is set up as a binary classification task where tweets are assigned labels from the set \{\textit{ironic}, \textit{non-ironic}\}. A total of $4,024$ tweets were released by organizers as training data. In addition, a total of $1,006$ tweets were used by organizers as TEST data. TEST labels were not release; and teams were expected to submit the predictions produced by their systems on the TEST split. For our models, we split the $4,024$ released training data into 90\% TRAIN ($n=3,621$ tweets; `ironic'$=1,882$ and `non-ironic'$=1,739$) and 10\% DEV ($n=403$ tweets; `ironic'$=209$ and `non-ironic'$=194$). We use the same small-GRU architecture of Section 3.1 as our baselines. We fine-tune BERT on our TRAIN, and evaluate on DEV. The small-GRU obtain 73.70\% acc. and 73.47\% $F_1$ score. BERT model significantly outperforms the small-GRU, acquiring 81.64\% acc. and 81.62\% $F_1$ score. 

\begin{table}[h]
\centering
\begin{tabular}{l|r|r}
\hline
              & \multicolumn{1}{c|}{\textbf{Acc}} & \multicolumn{1}{c}{\textbf{F}$_1$} \\ \hline
\textbf{small-GRU}  & 73.70                             & 73.47                            \\ 
\textbf{BERT} & \textbf{81.64}                             & \textbf{81.62}                            \\ \hline
\end{tabular}\caption{Model performance on irony detection.}
\end{table}

\subsection{Sentiment}\label{subsec:senti}
We collect 15 datasets related to sentiment analysis of Arabic, including MSA and dialects~\cite{abdul2012awatif,abdulla2013arabic,abdul2014samar,nabil2015astd,kiritchenko2016semeval,aly2013labr,salameh2015sentiment,rosenthal2017semeval,alomari2017arabic,mohammad2018semeval,baly2019arsentd}. Table~\ref{tab:sentiment} shows all the corpora we use. The datasets involve different types of sentiment analysis tasks such as binary classification (i.e., negative or positive), 3-way classification (i.e., negative, neutral, or positive), and subjective language detection. To combine these datasets for binary sentiment classification, we normalize different types of labels to binary tags in the set $\{`positive', `negative'\}$ using the following rules:
\begin{itemize}
  \item Map \{Positive, Pos, or High-Pos\} to `positive'
  \item Map \{Negative, Neg, or High-Neg\} to `negative'
  \item Exclude samples whose label is not `positive' or `negative' such as `obj', `mixed', `neut', or `neutral'.
\end{itemize}

After label normalization, we obtain $126,766$ samples. We split this resulting dataset into 80\% training (TRAIN), 10\% development (DEV), and 10\% test (TEST). The distribution of classes in our splits is presented in Table~\ref{tab:sentiment_data}. We fine-tune pre-trained BERT on the TRAIN set using PyTorch implementation with $2e-6$ learning rate and 15 epochs, as explained in Section~\ref{sec:methods}. Our best model on the DEV set obtains 80.24\% acc. and 80.24\% $F_1$. We evaluate this best model on TEST set and obtain 77.31\% acc. and 76.67\% $F_1$. 

\begin{table}[h]
\centering
\begin{tabular}{l|r|r|r}
\hline
                  & \textbf{TRAIN} & \textbf{DEV} & \textbf{TEST} \\ \hline
\textbf{\# pos} & 61,555          & 7,030         & 7,312          \\ 
\textbf{\# neg} & 39,044          & 7,314         & 4,511          \\ 
\textbf{Total}    & 100,599         & 14,344        & 11,823         \\ \hline
\end{tabular}\caption{Distribution of sentiment classes in our data splits.}\label{tab:sentiment_data}
\end{table}

\footnotesize
\begin{center}
\begin{table*}[h]
\begin{tabular}   {|p{3.8cm}|c|p{2 cm}|p{1cm}|c|p{3.5cm}|c|}

\hline
 \textbf{Authors }& \textbf{Task} & \textbf{Sources}&  \textbf{\# Data}  &\textbf{ \#Class } &\textbf{ Classes }    &\textbf{MSA/DIA} \\
\hline

\hline
~\newcite{abdul2012awatif} &   SSA & Wiki.\textbf{\footnote{\textcolor[rgb]{0,0,1}{\url{https://www.wikipedia.com}}}}, PAT\textbf{\footnote{\textcolor[rgb]{0,0,1}{\url{https://catalog.ldc.upenn.edu/LDC2005T20}}}}, Forums & $5,382$ &   4 & Obj,
Subj, Pos, Neg and Neut  & MSA  \\ 
~\newcite{abdulla2013arabic} &    SA & Twitter,  & $2000$ &   2& Pos, Neg & MSA \\ 
~\newcite{abdul2014samar} & SSA & Maktoob\textbf{\footnote{\textcolor[rgb]{0,0,1}{\url{chat.mymaktoob.com}}}}, Twitter & 11918 & 3 &   Obj,
Subj Pos, Subj Neg and Subj Mixed  & MSA+DIA  \\ 
~\newcite{nabil2015astd} &    SSA & Twitter & $10000$ &   4 & Obj, Subj Pos, Subj Neg and Subj Mixed  & MSA \\ 
~\newcite{kiritchenko2016semeval}  &   SI &Twitter & $1,366$ & $-$ & Regression [0,1] & MSA\\ 
~\newcite{aly2013labr} &    SA & Book reviews & $63,000$ &   3& Pos, Neg, or Neut & MSA\\ 
~\newcite{salameh2015sentiment} &      SA   & BBN Parallel Text\textbf{\footnote{\textcolor[rgb]{0,0,1}{\url{ https://catalog.ldc.upenn.edu/LDC2012T09}}}}  & 1200 &   3 & Pos, Neg, or Neut & DIA     \\ 
~\newcite{salameh2015sentiment} &      SA  & Twitter   & 2000 &   3 &  Pos, Neg, or Neut   & DIA  \\ 
~\newcite{rosenthal2017semeval} &  SA &   Twitter&  9,500  &   2 &  Pos, or Neg  & MSA\\ 
~\newcite{rosenthal2017semeval} &   SA & Twitter &  3,400  &   3 & Pos, Neut, or Neg & MSA\\ 
~\newcite{rosenthal2017semeval} & SA & Twitter &  9,450  &  5  & High-Pos, Pos, Neut, Neg, Hihg-Neg & MSA\\ 
~\newcite{alomari2017arabic} &     SA & Twitter   & 1800 &   3 &Pos or Neg  & DIA    \\ 
~\newcite{mohammad2018semeval} &  SA & Twitter & 1,800 &   7 & Various levels of Pos, Neg or Neut [-3,3]  & MSA\\ 
Saad (2019)\textsuperscript{*}&   SA & Twitter &   58,751 &  2 &Pos, or Neg   & DIA\\ 
~\newcite{baly2019arsentd} &   SA &Twitter &  4,000 &   5& High-Pos, Pos, Neut, Neg, Hihg-Neg  & DIA\\ \hline
\end{tabular}
\caption{Sentiment analysis datasets. \textbf{SA}: Sentiment analysis.  \textbf{SSA}: Subjectivity and sentiment analysis. \textsuperscript{*}Dataaet from Saad (2019) is available at \url{ https://www.kaggle.com/mksaad/arabic-sentiment-twitter-corpus}.}
\label{tab:sentiment}
\end{table*}
\end{center}

\section{\sys Design and Use}\label{sec:design_use}
\sys consists of identifier tools including age, gender, dialect, emotion, irony and sentiment. Each tool comes with an embedded model. The tool comes with modules for performing normalization and tokenization. AraNet can be used either as (1) a Python library or (2) a command-line and interactive tool, as follows:

\paragraph{}

\textbf{\sys as a Python Library:} Importing AraNet module as a Python library provides identifier functions. Prediction is based on a text input or a path to a file, and returns the identified class label. The library also returns the probability distribution over all available class labels if needed. This probability is the outcome of the softmax function applied to the last layer (with logits) in each model. Figure~\ref{fig:uasge1} shows two examples of using the tool as Python library.

\paragraph{}

\textbf{\sys as a Command-Line and Interactive Tool:} AraNet provides scripts supporting both command-line and interactive mode. Command-line mode accepts a text or file path. Interaction mode is good for quick interactive line-by-line experiments and also pipeline re-directions. 

\begin{figure}[h]
\begin{centering}
 \frame{\includegraphics[width=75mm,scale=0.90]{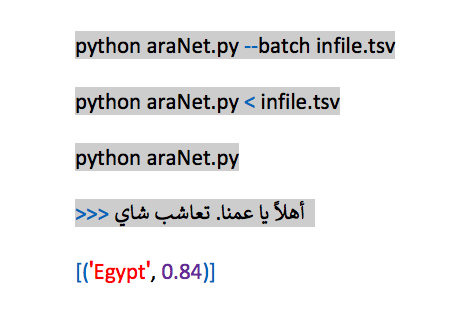}}
  \caption{AraNet usage examples as command-line mode, pipeline, and interactive mode.}
  \label{fig:uasge2}
  \end{centering}
\end{figure}

AraNet is available through pip or from
source on GitHub~\footnote{\url{https://github.com/UBC-NLP/aranet}} with detailed documentation.

\section{Ethical Considerations}~\label{sec:ethics}
AraNet is trained on data collected from publicly available sources. The distribution of classes across the different tasks are reasonably balanced as listed in the respective sections in the current paper. Meanwhile, we note that we have not used AraNet in real-world situations, nor tested any bias its decisions could involve. As a result, we advise against using AraNet in decision making without prior research as to what its deployment could involve and how best it can be tested. \textbf{\textit{We also do not approve any use of the AraNet or its decisions in any form for manipulative, unfair, malicious, dangerous, or otherwise unlawful (including by international standards) causes by individuals or organizations.}} Our conviction is that machine-learning-based software can be very powerful and useful, if not at times necessary, but must be tested and deployed \textit{only} carefully and ethically. AraNet is no exception.

\section{Related Works}\label{sec:rel_works}
As we pointed out earlier, there are several works on some of the tasks but less on others. By far, Arabic sentiment analysis has been the most popular task. Works focused on both MSA~\cite{mageedranlp2011,mageed2014samar} and dialects~\cite{nabil2015astd,elsahar2015building,al2015deep,al2018arabic,al2019using,al2019comprehensive,farha2019mazajak}. A number of studies have been published on dialect detection, including~\cite{zaidan2011arabic,zaidan2014arabic,elfardy2013sentence,cotterell2014multi}. 
Some works took as their target the tasks of age detection~\cite{zaghouani2018arap,rangel2019ADPA}, gender detection~\cite{zaghouani2018arap,rangel2019ADPA}, irony identification~\cite{karoui2017soukhria,idat2019}, and emotion analysis~\cite{abdul2016dina,alhuzali2018enabling}.

\paragraph{}

A number of resources and tools exist for Arabic natural language processing, including Penn Arabic treebank~\cite{maamouri2004penn}, Buckwalter Morphological Analyzer~\cite{buckwalter2002buckwalter}, segmenters~\cite{abdelali2016farasa}, POS taggers~\cite{abumalloh2016arabic,diab2004automatic}, morpho-syntactic analyzers~\cite{abdul2013asma,pasha2014madamira}, subjectivity and sentiment analysis~\cite{mageed2019modeling,farha2019mazajak}, offensive and hateful language~\cite{elmadani_osact4}, and dangerous speech~\cite{alshehri_osact4}.

\section{Conclusion}~\label{sec:conc}
We presented AraNet, a deep learning toolkit for a host of Arabic social media processing. \sys predicts age, dialect, gender, emotion, irony, and sentiment from social media posts. It delivers either state-of-the-art or competitive performance on these tasks. It also has the advantage of using a unified, simple framework based on the recently-developed BERT model. \sys has the potential to alleviate issues related to comparing across different Arabic social media NLP tasks, by providing one way to test new models against \sys predictions (i.e., model-based comparisons). Our toolkit can be used to make important discoveries about the Arab world, a vast geographical region of strategic importance. It can enhance also enhance our understating of Arabic online communities, and the Arabic digital culture in general. 

\section{Bibliographic References}
\bibliographystyle{lrec}
\bibliography{lrec2020}

\end{document}